\documentclass[journal,twoside,web]{ieeecolor}
\usepackage{jsen}
\usepackage{cite}
\usepackage{amsmath,amssymb,amsfonts}
\usepackage{graphicx}
\usepackage{textcomp}
\usepackage{booktabs}
\usepackage{array}
\usepackage{siunitx}

\def\BibTeX{{\rm B\kern-.05em{\sc i\kern-.025em b}\kern-.08em
    T\kern-.1667em\lower.7ex\hbox{E}\kern-.125emX}}
\definecolor{abstractbg}{rgb}{0.89804,0.94510,0.83137}
\begin{document}

\title{Aligned Radiometric RGB--Thermal Fusion for UAV Facade Anomaly Screening}

\author{Yuan Yang, Shulei Li, and Haobo Liang
\thanks{Yuan Yang and Haobo Liang are with the Hong Kong Center for Construction
Robotics, Hong Kong SAR, China (e-mail: yuanyang@ust.hk; hbliang@ust.hk).}
\thanks{Shulei Li is with the Hong Kong Center for Construction Robotics, Hong Kong
SAR, China (e-mail: chelsea616.li@connect.polyu.hk).}
\thanks{Corresponding author: Haobo Liang.}}

\IEEEtitleabstractindextext{%

\begin{abstract}
Unmanned aerial vehicle facade inspection can combine red, green, and blue (RGB) imagery with thermal measurements to screen surface and subsurface anomalies. However, geometric discrepancies between the sensors and thermal image rendering can obscure spatial correspondence and weak temperature contrasts. This article presents a sensor-level pipeline comprising per-sensor correction, RGB-to-thermal registration, common-support cropping, and signed local contrast encoding of 16-bit radiometric measurements. The encoding preserves the distinction between locally hotter and colder regions and supplies the fourth input channel of a compact single-stream detector. We introduce M3T, a dataset of 674 paired RGB and radiometric thermal samples from five facade-inspection projects covering eight component and anomaly categories. The median residual registration error is 3.384 pixels, and a controlled-displacement analysis characterizes how the local contrast response changes under controlled displacement. Project-grouped four-fold evaluation yields mean average precision of 0.168 over intersection-over-union thresholds from 0.5 to 0.95, using 28.50 billion floating-point operations per image. A separate single-split ablation shows improved delamination detection over RGB-only and alternative thermal inputs, although aggregate accuracy does not improve over RGB alone. Evaluation on RGBT-Tiny shows mixed performance with rendered thermal imagery. These results characterize the category-specific benefits and limitations of aligned radiometric contrast for compact facade screening.

\end{abstract}

\begin{IEEEkeywords}
Facade inspection, multisensor alignment, radiometric thermal imaging,
RGB--thermal fusion, unmanned aerial vehicles.
\end{IEEEkeywords}}

\maketitle

\section{Introduction}
\label{sec:introduction}

\IEEEPARstart{B}{uilding} facade inspection requires the identification of
surface deterioration, subsurface anomalies, and facade components over large
and often difficult-to-access areas. Unmanned aerial vehicles (UAVs) provide a
flexible means of acquiring close-range facade imagery without extensive
scaffolding or fixed sensing infrastructure~\cite{Fan2025UAVFacade,
Li2024Debonding}. When visible and thermal cameras are used together, RGB
images describe texture and structural boundaries, while thermal measurements
record spatial variations in surface temperature that may indicate hidden or
weakly visible conditions. Such measurements are therefore useful for
screening candidate regions before detailed inspection, rather than replacing
the final assessment performed by an inspector.

A paired RGB--thermal acquisition is not immediately suitable for multimodal
detection. Consumer UAV cameras use separate optical paths, resolutions, and
fields of view, producing geometric discrepancies between their measurements.
Direct concatenation can consequently associate an RGB structure with an
incorrect thermal location or retain borders observed by only one sensor.
Thermal representation introduces a second difficulty. Automatic gain control
and global normalization commonly convert high-dynamic-range measurements into
8-bit images for visualization, but weak local temperature differences can be
compressed during this process~\cite{Javidnia2025AGC,Gil2024Fieldscale,
Lee2024ThermalChameleon}. Moreover, thermographic visibility depends on surface
material, thermal excitation, and ambient conditions~\cite{Balaras2002IRT,
Kylili2014IRT,Mahmoodzadeh2023AerialThermography}. Consequently, an anomaly representation should not assume a fixed thermal
polarity or magnitude across projects. It should preserve both positive
and negative local deviations while limiting dependence on the scene-wide
temperature range. Geometry and radiometric
representation must therefore be addressed together when constructing
detector inputs.

Existing RGB--thermal benchmarks primarily address pedestrian, vehicle, and
general-scene perception using rendered infrared imagery, including KAIST,
LLVIP, M3FD, and RGBT-Tiny~\cite{Hwang2015KAIST,Jia2021LLVIP,
Liu2022TarDAL,Ying2025RGBTTiny}. Corresponding detection methods have developed
increasingly sophisticated cross-modal interaction mechanisms, such as
calibrated transformer fusion and iterative dual-stream
attention~\cite{Yuan2024C2Former,Shen2024ICAFusion}. These developments provide
strong architectural baselines, but facade thermography also requires
questions to be resolved at the sensor level: whether the two measurements
share valid spatial support, how much radiometric response remains under
registration error, and whether local contrast encoding preserves weak anomaly
signals. Public paired datasets that retain radiometric measurements for UAV
facade anomaly detection also remain uncommon. Without such data and intermediate sensor-level measurements, the effects
introduced by calibration, radiometric encoding, and detector architecture
are difficult to distinguish.

This paper presents a sensing-to-detection pipeline built around aligned radiometric measurements. Both sensors are corrected independently, and the RGB observation is mapped into the thermal coordinate frame before their common image support is retained. A signed local radiometric contrast representation, denoted by $C_2$, encodes positive and negative deviations from a locally estimated temperature background as the fourth channel of a compact single-stream detector. We examine the resulting inputs through registration measurements and encoding ablations, and evaluate downstream detection across project-grouped folds. This evaluation assesses where radiometric contrast contributes useful anomaly cues and where its benefit remains limited.

The main contributions are summarized as follows:
\begin{enumerate}
\item We construct the M3T facade dataset using a DJI Mavic 3 Thermal platform.
It contains 674 paired RGB--radiometric thermal samples from five inspection
projects, with bounding-box annotations for eight facade-element and anomaly
categories. The original radiometric thermal files, RGB images, annotations,
and project-grouped splits will be released upon publication. To our knowledge, upon release, M3T will be one of the few publicly
available UAV facade inspection datasets that retain 16-bit radiometric
thermal measurements.

\item We develop a sensor-level sample-construction pipeline comprising
per-sensor correction, RGB-to-thermal registration, common-support cropping,
and signed local radiometric contrast encoding. The $C_2$ representation
adapts locality-aware radiometric processing to preserve locally hotter and
colder responses within a four-channel detector input, without introducing
dual-stream branches or custom cross-modal interaction modules.

\item We characterize the sensing-to-detection chain through landmark-based registration error, controlled-displacement contrast response, and project-grouped four-fold detection. A separate single-split ablation compares RGB-only input with three thermal encodings to assess their category-specific contributions. Comparisons with EME, ICAFusion, and CMA-Det, together with complexity profiling and evaluation on RGBT-Tiny, quantify the accuracy–complexity tradeoff and examine performance beyond the radiometric facade setting.

\end{enumerate}

\section{Related Work}
\label{sec:related}

\subsection{RGB-Thermal Fusion for Object Detection}
\label{sec:related_fusion}

RGB-thermal object detection has been studied extensively for pedestrian,
automotive, and general multispectral perception. Paired benchmarks such as
KAIST~\cite{Hwang2015KAIST}, LLVIP~\cite{Jia2021LLVIP}, and
M3FD~\cite{Liu2022TarDAL} established widely used visible-infrared evaluation
settings. More recently, RGBT-Tiny~\cite{Ying2025RGBTTiny} extended this
literature to paired visible-thermal tiny-object detection and introduced a
scale-adaptive fitness measure. Existing fusion architectures are commonly
distinguished by where the modalities interact, including input-level fusion,
intermediate feature fusion, and prediction-level combination. Representative
approaches use cross-modal feature exchange~\cite{Jang2025MCOR}, guided
attentive fusion~\cite{Zhang2021GAFF}, and strategies that explicitly address
modality imbalance in multispectral detection~\cite{Kim2024CMM}.

Recent work has pursued stronger cross-modal interaction through calibrated
and complementary transformer features~\cite{Yuan2024C2Former} and iterative
dual cross-attention in ICAFusion~\cite{Shen2024ICAFusion}. DAMSDet
combines competitive modality-aware query selection with deformable
multispectral feature aggregation~\cite{Guo2024DAMSDet}. Other methods
investigate wavelet-based Mamba fusion~\cite{Zhu2025WaveMamba}, state-space
modeling of shared and cross-modal representations~\cite{Shen2026MS2Fusion},
and frequency-decoupled cross-modal learning for tiny
objects~\cite{Li2026DyFCLT}. These designs demonstrate the breadth of
architectural strategies available for visible-infrared detection, but their
computational costs, training requirements, and sensitivity to modality
quality differ across application settings.

A separate line revisits efficient input-level fusion.
Zhang \emph{et al.} identify information interference in naive early fusion
and improve a single-branch detector using a shape-priority fusion strategy,
weak supervision, and knowledge
distillation~\cite{Zhang2025EarlyFusion}. This result does not imply that
simple channel concatenation is universally sufficient. Rather, it
demonstrates that early fusion requires deliberate treatment of modality
interference. Condition-aware UAV
benchmarks show that modality utility depends on acquisition
conditions~\cite{Chen2025UAVFusion}. Our work builds on this perspective by resolving cross-modal geometry and
constructing a radiometric contrast representation before input-level fusion
in a simple four-channel single-stream detector.

\subsection{Cross-Modal Alignment and Misalignment-Robust Detection}
\label{sec:related_alignment}

Spatial inconsistency between visible and thermal images is a recurring
challenge in multispectral detection. Weakly aligned cross-modal learning
addresses correspondence errors through region-level
reasoning~\cite{Zhang2019WeaklyAligned}. Subsequent work investigates
adaptive dual-discrepancy calibration~\cite{He2023ADCNet} and feature
alignment for multimodal UAV imagery~\cite{Chen2024OAFA}. CF-Deformable DETR
uses cross-modal deformable attention to learn point correspondences without
explicit image registration~\cite{Fu2024CFDETR}, whereas YOLO-Adaptor studies
an adaptive one-stage architecture for non-aligned
inputs~\cite{Fu2024YOLOAdaptor}. COMO combines cross-Mamba interaction with
offset-guided feature fusion to address multimodal feature
misalignment~\cite{Liu2026COMO}.

These developments are supported by drone-based datasets that emphasize
different sensing conditions. DroneVehicle targets RGB-infrared vehicle
detection from aerial imagery~\cite{Sun2022DroneVehicle}, while DVTOD
explicitly evaluates misaligned visible-thermal drone
imagery~\cite{Song2024DVTOD}. Related RGB-X semantic segmentation
frameworks, including CMX~\cite{Liu2023CMX} and
CMNeXt~\cite{Zhang2023DeLiVER}, use cross-modal rectification and feature
interaction, although their task is dense semantic segmentation rather than
box-level detection.

Many of these approaches estimate or accommodate correspondence errors
within the detection network. Such mechanisms can address a range of
misalignment conditions, but they do not replace an explicit analysis of
sensor resolution asymmetry, common image support, and the effect of
registration error on a weak radiometric anomaly. In our setting, the visible
and thermal sensors provide substantially different native resolutions and
fields of view. We therefore first establish a shared valid image region
through sensor-level registration and then quantify both alignment accuracy
and the degradation of anomaly contrast under controlled spatial
perturbations. This sensor-level preparation complements feature-level
alignment rather than establishing that existing alignment-aware detectors
are limited to small displacements.

\subsection{UAV Thermography for Building Facade Inspection}
\label{sec:related_thermography}

Infrared thermography is an established nondestructive technique for building
diagnostics~\cite{Balaras2002IRT,Kylili2014IRT}. Its combination with UAV
imaging has been studied for facade
debonding~\cite{Li2024Debonding}, heat-loss
inspection~\cite{Waqas2024HeatLoss}, building-envelope thermal anomaly
analysis~\cite{Mirzabeigi2025Buildings}, automated facade
inspection~\cite{Fan2025UAVFacade}, and three-dimensional thermal
inspection~\cite{Lin2025True3D}. Direct comparisons of multimodal fusion
strategies for exterior wall defects further demonstrate the relevance of
visible-thermal sensing to this application~\cite{Yang2023RGBThermalFacade}.
More recently, weakly aligned cross-modal learning has been used for
subsurface facade defect segmentation~\cite{He2025WCL}, and broader UAV
datasets have increased the diversity of available sensing modalities and
imaging conditions~\cite{Wang2025UAVScenes,Chen2025UAVFusion}.

Despite these advances, quantitative aerial thermography remains sensitive to
surface emissivity, solar loading, recent weather, acquisition time, and
thermal sensor stability~\cite{Balaras2002IRT,
Mahmoodzadeh2023AerialThermography,Wang2023Drift}. Consequently, a local
thermal anomaly associated with facade delamination or moisture can appear
prominent in one acquisition and weak or ambiguous in another. Thermal
contrast is therefore better treated as condition-dependent evidence than as
an equally reliable cue for every defect class.

Our application differs from studies that rely on thermal images alone or
produce only dense segmentation outputs. We consider aligned RGB and
radiometric thermal measurements from a consumer UAV and evaluate a common
four-channel detector across visible facade elements and thermally expressed
anomalies. The resulting objective is human-in-the-loop candidate screening,
not autonomous diagnosis or a claim that thermal fusion uniformly improves
aggregate detection accuracy.

\subsection{Radiometric Thermal Representation}
\label{sec:related_radiometric}

The representation of thermal measurements before downstream processing can
strongly influence the visibility of weak anomalies. Fieldscale adapts
thermal image rescaling to local scene
structure~\cite{Gil2024Fieldscale}, while Thermal Chameleon investigates
task-adaptive tone mapping for radiometric thermal-infrared
images~\cite{Lee2024ThermalChameleon}. Task-driven adaptive gain control
likewise studies 16-bit-to-8-bit thermal conversion for object
detection~\cite{Javidnia2025AGC}. Multimodal acquisition systems such as MM5
preserve raw thermal measurements together with other sensing
modalities~\cite{Brenner2026MM5}, and corrections for temporal drift have
been investigated for uncooled UAV thermal
imagers~\cite{Wang2023Drift}.

Our signed local contrast channel shares the motivation of this literature
but serves a different role. Rather than generating a visually enhanced
thermal image or introducing a new tone-mapping operator, we subtract a
locally estimated radiometric background and symmetrically normalize the
result so that positive and negative deviations remain distinguishable.
This representation is constructed as an aligned detector input and is
evaluated through class-specific contrast response, registration sensitivity,
and downstream anomaly screening. Its contribution lies in connecting
radiometric preprocessing to measurable sensing behavior in a weak-signal
facade setting, rather than claiming a novel image-processing operation.

\section{Sensor-Level Processing and Anomaly Screening}
\label{sec:method}

\begin{figure*}[t]
\centering
\includegraphics[width=\textwidth]{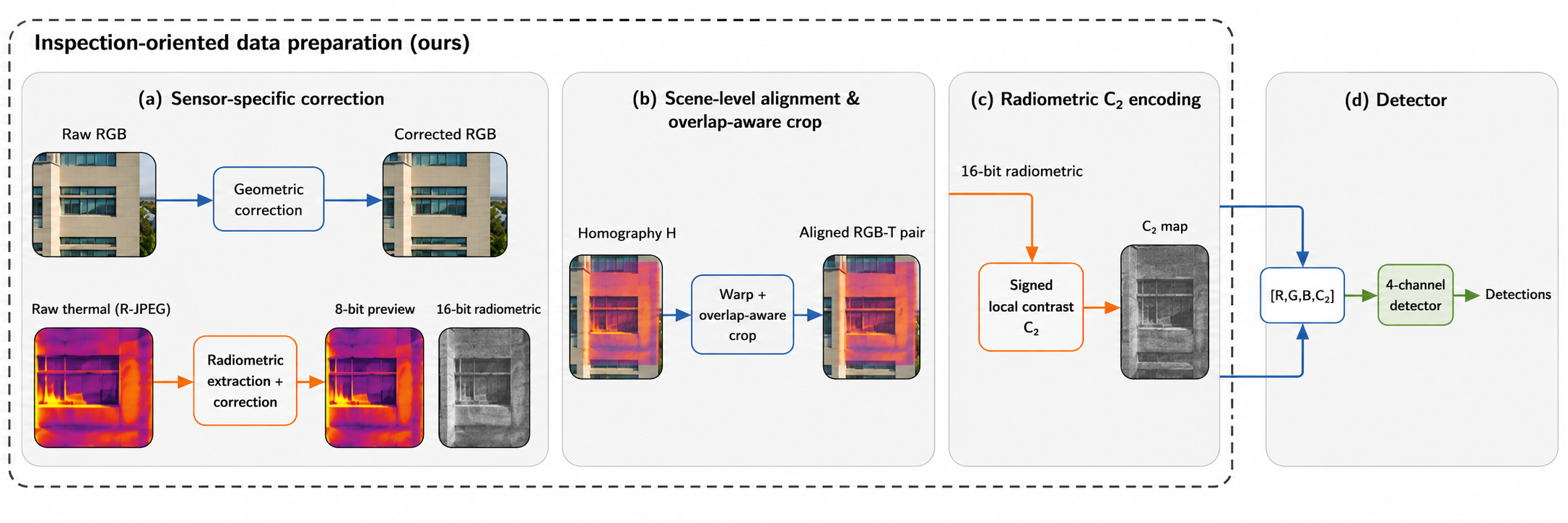}
\caption{Overview of the sensor-level processing pipeline. Raw RGB and
R-JPEG thermal captures are first corrected for lens distortion using
per-sensor checkerboard calibration, and per-pixel radiometric temperatures
are extracted as 16-bit data. A scene-level homography
$\mathbf{H}_{\mathrm{r}\rightarrow\mathrm{t}}$ maps the corrected RGB image
into the thermal coordinate frame, after which common-support cropping
produces aligned RGB--thermal pairs. The corrected radiometric measurements
are then encoded as the signed local radiometric contrast channel $C_2$.
The resulting four-channel input $\mathbf{X}=[R,G,B,C_2]$ is processed by a
single-stream detector. The dashed frame denotes the sensor-processing and
sample-construction stages applied before detection.}
\label{fig:pipeline}
\end{figure*}

Commodity RGB-thermal UAV platforms present three coupled difficulties for
facade inspection: severe resolution asymmetry between the two sensors, the
absence of native pixel-level correspondence, and weak, condition-dependent
thermal signatures. We address these at the sensor and sample-construction
level rather than inside the detector. As summarized in
Fig.~\ref{fig:pipeline}, the pipeline corrects each sensor and extracts
radiometric temperature, recovers a scene-level homography to produce
overlap-aware aligned crops, and encodes a signed local radiometric contrast
channel that supplies the fourth detector input.

Two constraints organize this design. First, cross-modal geometry is resolved
before training, so that every detector, including the fusion baselines used
later for comparison, receives samples on the same valid support. Second,
radiometric information is kept in temperature space until the final encoding
step, avoiding premature 8-bit compression that would attenuate the weak local
contrasts relevant to facade anomalies. How the value of the resulting channel
varies across anomaly categories is examined in the experimental analysis.

\subsection{Sensor Correction and Radiometric Extraction}

Let $I_{\mathrm{rgb}}^{\mathrm{raw}}$ and
$I_{\mathrm{th}}^{\mathrm{raw}}$ denote the raw RGB image and the thermal
R-JPEG acquired by the DJI Mavic 3 Thermal. The two sensors have substantially
different native resolutions, namely $4000\times3000$ for RGB and
$640\times512$ for thermal imaging, as well as different fields of view and
lens distortion characteristics. The RGB and thermal cameras are therefore
calibrated separately using checkerboard observations. The estimated camera
intrinsics and radial distortion coefficients are applied to undistort each
modality and remove invalid image borders, following the dual-sensor
calibration procedures used in UAV multimodal
inspection~\cite{Fan2025UAVFacade,Lin2025True3D}.

The thermal R-JPEG contains an 8-bit preview together with embedded
radiometric measurements. We decode the radiometric payload using the DJI
Thermal SDK and store the resulting per-pixel measurements as a 16-bit
temperature array, denoted by $T_{16}$. The same geometric correction and
cropping applied to the thermal preview are also applied to $T_{16}$ so that
their pixel coordinates remain consistent. The corrected 8-bit preview is
used only to establish visible cross-modal correspondences during alignment,
whereas the signed local contrast channel introduced in
Section~\ref{sec:method} is calculated directly from $T_{16}$. No display
tone mapping or global 8-bit normalization is applied to the radiometric
measurements before this encoding step.

\subsection{RGB-to-Thermal Registration and Common-Support Cropping}

After individual sensor correction, the RGB and thermal outputs remain
spatially inconsistent because of their different resolutions, viewpoints,
and fields of view. We estimate a scene-level homography
$\mathbf{H}_{\mathrm{r}\rightarrow\mathrm{t}}$ that maps coordinates from the
corrected RGB image to the corrected thermal image plane. For a homogeneous
RGB coordinate $\widetilde{\mathbf{p}}_{\mathrm{rgb}}$, its corresponding
thermal coordinate is given by
\begin{equation}
\widetilde{\mathbf{p}}_{\mathrm{th}}
\sim
\mathbf{H}_{\mathrm{r}\rightarrow\mathrm{t}}
\widetilde{\mathbf{p}}_{\mathrm{rgb}} .
\label{eq:rgb_to_thermal}
\end{equation}

The transformation is determined by overlaying the corrected RGB image and
thermal preview and interactively adjusting translation, rotation, and scale
until common facade structures, including window boundaries, vents, and
air-conditioning units, are aligned. Nearby images acquired within the same
timestamp-based inspection group share similar sensor geometry and facade
viewpoints. A single transformation is therefore estimated for each group
and reused for its constituent image pairs.

The corrected RGB image is warped into the thermal coordinate frame:
\begin{equation}
I_{\mathrm{rgb}}^{\mathrm{align}}
=
\mathcal{W}\!\left(
I_{\mathrm{rgb}}^{\mathrm{corr}},
\mathbf{H}_{\mathrm{r}\rightarrow\mathrm{t}}
\right),
\label{eq:rgb_warp}
\end{equation}
where $\mathcal{W}(\cdot)$ denotes perspective warping. The radiometric
temperature map remains in the corrected thermal coordinate system, avoiding
an additional resampling of the temperature measurements during cross-modal
registration.

A common-support mask is obtained from the intersection between the valid
warped RGB region and the valid thermal image region. Non-overlapping borders
are removed using the bounding extent of this mask. Bounding-box annotations
are transformed by $\mathbf{H}_{\mathrm{r}\rightarrow\mathrm{t}}$, shifted
according to the crop origin, and clipped to the retained image support. The
resulting sample consists of an aligned RGB crop, its corresponding
radiometric temperature map, and annotations defined in the same thermal
coordinate system. Registration accuracy and sensitivity to residual spatial
displacement are evaluated in
Section~\ref{sec:registration_evaluation}.

\subsection{Signed Local Radiometric Contrast Encoding}

Most RGB-thermal detectors operate on rendered 8-bit thermal frames. We instead
derive the fourth input channel from the corrected 16-bit radiometric
measurements, following the motivation of locality-aware thermal processing and
task-adaptive radiometric
mapping~\cite{Javidnia2025AGC,Gil2024Fieldscale,Lee2024ThermalChameleon}, but
using the result as a fusion channel rather than for visual enhancement.

Let $T_{16}(x)$ denote the corrected radiometric value at pixel $x$, expressed
in degrees Celsius and decoded from the 16-bit radiometric payload. A locally
estimated background field is obtained by Gaussian smoothing:
\begin{equation}
B(x) = (G_\sigma * T_{16})(x),
\label{eq:background}
\end{equation}
where $G_\sigma$ is a Gaussian kernel with $\sigma = 11\,\mathrm{px}$. The local
thermal deviation is
\begin{equation}
\Delta T(x) = T_{16}(x) - B(x).
\label{eq:deltat}
\end{equation}
Positive values indicate locally hotter regions and negative values locally
colder regions. To obtain a fusion-ready single-channel representation, we apply
symmetric clipped normalization:
\begin{equation}
C_2(x) = \mathrm{clip}\!\left(\frac{\Delta T(x)}{2\,p_{\mathrm{clip}}}
+ \frac{1}{2},\; 0,\; 1\right),
\label{eq:c2}
\end{equation}
where $p_{\mathrm{clip}} = \max\!\big(\mathrm{percentile}_{99}(|\Delta T|),\,
\epsilon\big)$ sets an adaptive dynamic range, with $\epsilon = 0.05$ degrees Celsius. Neutral regions map to $0.5$,
while locally hot and cold anomalies are preserved on opposite sides of this
value; in implementation the normalized channel is mapped to the $[0,255]$
range so that it matches the numeric range of the RGB channels.

The detector input is
\begin{equation}
\mathbf{X} = [R, G, B, C_2],
\label{eq:input}
\end{equation}
where the first three channels are the aligned RGB crop and the fourth is the
$C_2$ map. We use a compact single-stage anchor-free detection backbone (approximately 11.1\,M parameters) with the input channel count changed from three to four; no dual-stream branches, custom
fusion layers, or cross-modal attention modules are added. HSV augmentation is
disabled for this four-channel input and its component ablations so that the
radiometric channel is not altered during training. The model is initialized by
pretraining on M3FD~\cite{Liu2022TarDAL} with an 8-bit infrared fourth channel
and then fine-tuned on the aligned M3T facade data with $C_2$, following the
input-level early-fusion detection strategy studied
in~\cite{Zhang2025EarlyFusion}.

\section{Experimental Evaluation}
\label{sec:experiments}

\subsection{Facade Dataset and Project-Grouped Evaluation}
\label{sec:facade_dataset}

The M3T facade dataset was collected using a DJI Mavic 3 Thermal UAV during
exterior building inspections. Each sample contains an RGB image, a
radiometric thermal measurement, and bounding-box annotations defined on the
aligned common image support described in Section~\ref{sec:method}. After
sensor correction, cross-modal registration, overlap-aware cropping, and
quality screening, the dataset contains 674 aligned RGB--thermal image pairs.
The annotations cover eight facade categories: surface crack, paint peeling,
efflorescence, window, air-conditioning unit, vent, delamination, and water
seepage.

The images originate from five inspection projects, denoted P1--P5.
To avoid distributing images from the same building environment across
training and validation sets, four project-level folds are formed by holding
out P1, P2, P3, and P4+P5, respectively. Their corresponding
training/validation image counts are 418/256, 587/87, 473/201, and 544/130.
In each fold, the remaining project groups are used for training.

Detection performance is measured using precision, recall, mean average
precision at an intersection-over-union threshold of 0.5
($\mathrm{mAP}_{50}$), and mean average precision averaged over thresholds
from 0.5 to 0.95 ($\mathrm{mAP}_{50:95}$). Results are reported for each
held-out project group and summarized using the mean and standard deviation
across the four folds. Per-category results are retained to characterize the
variation associated with the uneven class frequencies among projects.

\subsection{Registration Accuracy and Misalignment Sensitivity}
\label{sec:registration_evaluation}

Residual correspondence error is measured on the final aligned RGB--thermal
samples using manually selected structural landmarks. Seven facade scenes
are drawn from different acquisition sequences, with ten correspondence pairs
identified in each scene. The landmarks are placed on features visible in
both modalities, including window corners, facade boundaries, vents, and
air-conditioning units. The full-resolution RGB image is mapped into the
aligned thermal coordinate system using its stored homography. The residual
error of landmark $j$ is then calculated as
\begin{equation}
e_j =
\left\|
\mathbf{p}^{\mathrm{rgb}}_j -
\mathbf{p}^{\mathrm{th}}_j
\right\|_2 ,
\label{eq:registration_error}
\end{equation}
where $\mathbf{p}^{\mathrm{rgb}}_j$ and
$\mathbf{p}^{\mathrm{th}}_j$ denote the corresponding landmark coordinates
on the final aligned image support.

Across the 70 landmark pairs, the mean, median, and 95th-percentile
registration errors are 3.870, 3.384, and 7.434 pixels, respectively, as
summarized in Table~\ref{tab:sensor_level_results}(a). The scene-wise mean
errors range from 2.613 to 5.131 pixels.

The response to controlled displacement is evaluated on the P4+P5 held-out
fold, using 107 validation images that contain 159 delamination boxes and 102
water-seepage boxes. The instance counts therefore differ from the four-fold
totals in Table~\ref{tab:category_performance}.
The thermal measurement is shifted by 2, 5, 10, and 15 pixels in each
of the four cardinal directions, with border values replicated.
The C2 encoding is recomputed after each shift,
while the annotation boxes remain fixed.

For each box, the response is the mean of $|C_2(x)-0.5|$ over its
pixels, using $C_2$ normalized to $[0,1]$. Responses are averaged
over the four shift directions and then over boxes of the same
class, with equal weight assigned to each box. Denoting this
class-level mean by $A_c(d)$, the relative in-box amplitude is
\begin{equation}
R_c(d) =
\frac{A_c(d)}{A_c(0)} \times 100\%.
\label{eq:signal_retention}
\end{equation}
Here, $A_c(0)$ is computed from the unshifted input.

Table~\ref{tab:sensor_level_results}(b) shows that the delamination response
decreases to 97.2\% of its aligned value at 5 pixels and 77.0\% at 15 pixels.
The water-seepage response does not decrease over the tested offsets and stays
close to its aligned value. This statistic measures local contrast within fixed
boxes and does not distinguish anomaly-related responses from other thermal
structure. The water-seepage result therefore does not establish whether the
anomaly-specific signal is preserved under displacement.
The observed changes therefore characterize the representation's
response to displacement, without establishing detection
robustness under misalignment.

\begin{table}[t]
\centering
\caption{Sensor-level measurements: (a) residual registration error
and (b) relative in-box $C_2$ amplitude under controlled displacement.}
\label{tab:sensor_level_results}
\footnotesize
\setlength{\tabcolsep}{5pt}
\renewcommand{\arraystretch}{1.15}

\textbf{(a) Registration error (px)}\\[3pt]
\begin{tabular}{lccc}
\toprule
Set & Mean & Median & P95 \\
\midrule
Overall & 3.870 & 3.384 & 7.434 \\
\bottomrule
\end{tabular}

\vspace{6pt}

\textbf{(b) Relative in-box $C_2$ amplitude (\%)}\\[3pt]
\begin{tabular}{lccccc}
\toprule
Class & 0 px & 2 px & 5 px & 10 px & 15 px \\
\midrule
Delamination  & 100.0 & 99.6 & 97.2 & 88.4 & 77.0 \\
Water seepage & 100.0 & 100.2 & 100.6 & 101.5 & 103.5 \\
\bottomrule
\end{tabular}

\end{table}

\subsection{Project-Grouped Facade Detection}
\label{sec:project_grouped_detection}

We next evaluate downstream detection performance under the project-grouped
protocol described in Section~\ref{sec:facade_dataset}. The proposed four-channel detector is
compared with EME~\cite{Zhang2025EarlyFusion},
ICAFusion~\cite{Shen2024ICAFusion}, and
CMA-Det~\cite{Song2024DVTOD}. For each method, the official model
implementation and its corresponding multimodal input interface are retained.
All methods use the same aligned image pairs, category definitions, and
project-grouped training and validation partitions. Training is conducted at
an input resolution of $640\times640$ for a maximum of 150 epochs, with early
stopping retained for implementations that provide it. Each detector is
initialized from its corresponding M3FD pretrained checkpoint.

\begin{table*}[t]
\centering
\caption{Project-grouped M3T detection and complexity. Fold entries are
$\mathrm{mAP}_{50}/\mathrm{mAP}_{50:95}$; the last column reports the
four-fold mean $\pm$ sample standard deviation.}
\label{tab:project_detection}
\small
\setlength{\tabcolsep}{4pt}
\renewcommand{\arraystretch}{0.95}
\begin{tabular}{lccccccc}
\toprule
Method & Params (M) & GFLOPs & P1 & P2 & P3 & P4+P5 &
Mean $\pm$ SD \\
\midrule
EME
& 46.15 & 109.20
& 0.198/0.108
& 0.302/0.149
& 0.270/0.151
& 0.223/0.114
& $0.248\pm0.047$ / $0.131\pm0.023$ \\

ICAFusion
& 120.24 & 191.68
& \textbf{0.368/0.203}
& \textbf{0.407/0.212}
& \textbf{0.356/0.182}
& \textbf{0.395/0.165}
& $\mathbf{0.382\pm0.024}$ / $\mathbf{0.191\pm0.021}$ \\

CMA-Det
& 33.47 & 33.67
& 0.175/0.068
& 0.361/0.143
& 0.195/0.065
& 0.219/0.089
& $0.238\pm0.084$ / $0.091\pm0.036$ \\

Ours
& \textbf{11.13} & \textbf{28.50}
& 0.288/0.171
& 0.348/0.190
& 0.264/0.163
& 0.277/0.146
& $0.294\pm0.037$ / $0.168\pm0.018$ \\
\bottomrule
\end{tabular}
\end{table*}

Table~\ref{tab:project_detection} shows substantial variation across held-out
building projects. ICAFusion achieves the highest mean
performance, with an $\mathrm{mAP}_{50}$ of 0.382 and an
$\mathrm{mAP}_{50:95}$ of 0.191. The proposed four-channel detector obtains
the second-highest mean values, reaching 0.294 and 0.168, respectively. The
difference between the two methods is 0.088 at an IoU threshold of 0.5 and
decreases to 0.023 under the stricter $\mathrm{mAP}_{50:95}$ measure.

The proposed detector exceeds EME by 0.046 in $\mathrm{mAP}_{50}$ and 0.037
in $\mathrm{mAP}_{50:95}$. The corresponding differences relative to
CMA-Det are 0.057 and 0.077. Its standard deviations are 0.037 and 0.018 for
the two mAP measures, indicating limited variation in localization performance
across the four held-out project groups. CMA-Det exhibits the largest
fold-to-fold variation, particularly on the second group.

The two leading methods represent different accuracy--complexity operating
points. ICAFusion uses two modality-specific streams with iterative
cross-modal attention and achieves the highest mean detection accuracy. The
proposed model uses a single four-channel stream and obtains an
$\mathrm{mAP}_{50:95}$ of 0.168, compared with 0.191 for ICAFusion. Model
size, computational cost, and the corresponding detection accuracy are
examined jointly in the next subsection.

\subsection{Accuracy--Complexity and Category-Level Analysis}
\label{sec:accuracy_complexity}

\begin{figure}[t]
\centering
\includegraphics[width=\columnwidth]
{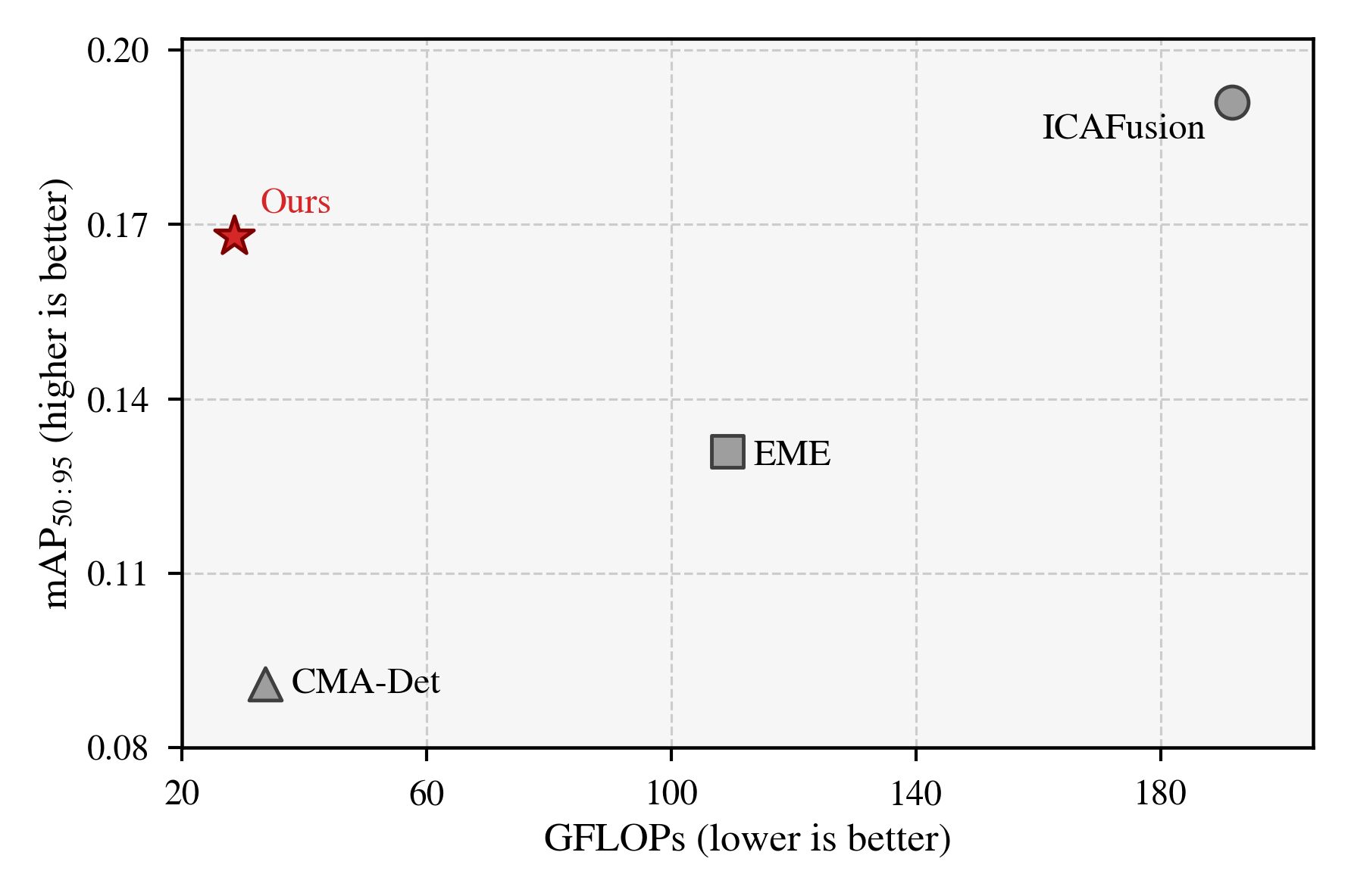}
\caption{M3T four-fold mean accuracy versus GFLOPs at $640\times640$.}
\label{fig:accuracy_efficiency}
\end{figure}

Table~\ref{tab:project_detection} and
Fig.~\ref{fig:accuracy_efficiency} jointly compare detection accuracy,
model size, and computational cost. Parameters are counted from the inference model
loaded by each repository, and computations are profiled at an input
resolution of $640\times640$ with a batch size of one. Both modality inputs
are supplied when profiling the dual-stream models, and one multiply--add is
counted as two floating-point operations.

ICAFusion provides the highest mean detection accuracy, reaching
$\mathrm{mAP}_{50}=0.382$ and
$\mathrm{mAP}_{50:95}=0.191$, with 120.24 M parameters and
191.68 GFLOPs. The proposed four-channel detector uses 11.13 M parameters
and 28.50 GFLOPs, corresponding to approximately 91\% fewer parameters and
85\% fewer floating-point operations. Its mean
$\mathrm{mAP}_{50:95}$ is 0.168, giving a difference of 0.023 from
ICAFusion under the stricter localization metric.

The proposed detector also uses fewer parameters and floating-point
operations than EME and CMA-Det while obtaining higher mean values on both
mAP measures. CMA-Det has a computational cost closer to that of the
proposed detector, but uses approximately three times as many parameters and
reaches 0.238 and 0.091 on the two mAP measures. These results establish two
different operating points: ICAFusion favors detection accuracy through
iterative dual-stream interaction, whereas the proposed model retains a
smaller single-stream inference graph after sensor-level alignment and
radiometric encoding.

Table~\ref{tab:category_performance} further reports the category-level
performance of the proposed detector. The instance count is accumulated over
the four held-out project groups. For a category absent from a particular
validation group, the corresponding fold is excluded from the category mean
rather than treated as having zero AP.

\begin{table}[t]
\centering
\caption{M3T category-level results of the proposed detector. Means use
only held-out folds containing the category.}
\label{tab:category_performance}
\footnotesize
\setlength{\tabcolsep}{3.5pt}
\renewcommand{\arraystretch}{0.92}
\begin{tabular}{lrrr}
\toprule
Category & Inst. & $\mathrm{AP}_{50}$ & $\mathrm{AP}_{50:95}$ \\
\midrule
Surface crack  & 1,578 & 0.017 & 0.0049 \\
Paint peeling  &   403 & 0.141 & 0.0639 \\
Efflorescence  &   109 & 0.003 & 0.0013 \\
Window         & 1,829 & 0.821 & 0.6293 \\
AC unit        &   199 & 0.526 & 0.2835 \\
Vent           &   435 & 0.629 & 0.2545 \\
Delamination   & 1,099 & 0.093 & 0.0289 \\
Water seepage  & 2,099 & 0.017 & 0.0043 \\
\bottomrule
\end{tabular}
\end{table}

The category-level results show that performance varies substantially across
facade elements and anomaly types. Window, AC unit, and vent have the
highest $\mathrm{AP}_{50}$ values, at 0.821, 0.526, and 0.629,
respectively. Their relatively well-defined geometry and visible boundaries
provide stable appearance cues across projects. In comparison, surface crack,
efflorescence, delamination, and water seepage remain difficult. Delamination
reaches an $\mathrm{AP}_{50}$ of 0.093, whereas water seepage reaches
0.017, consistent with its weaker and more spatially diffuse radiometric
response.

Paint peeling and efflorescence occur in only three of the four held-out
project groups, and efflorescence has the smallest overall instance count.
Their category-level estimates therefore reflect both limited support and
variation among buildings. Overall, the results indicate that the current
detector is more reliable for locating structurally distinct facade elements
than for resolving fine or environmentally dependent anomalies. The system is
therefore intended to identify candidate regions for subsequent review rather
than provide autonomous facade diagnosis.

\begin{figure*}[t]
\centering
\includegraphics[width=\textwidth]{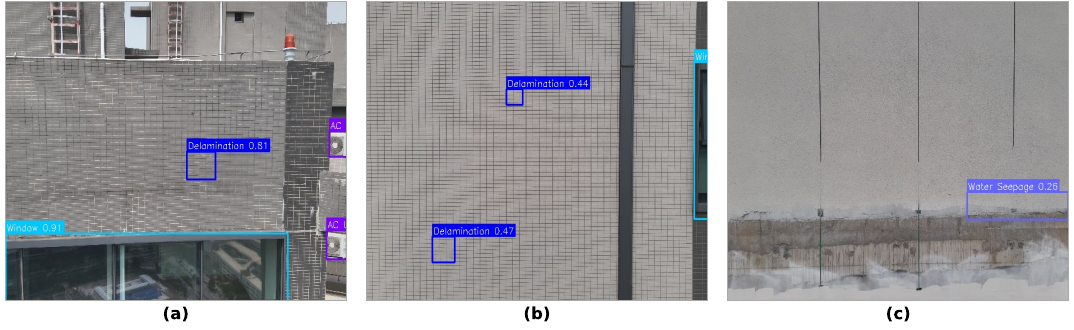}
\caption{Representative M3T validation detections: (a) and (b) facade
elements and delamination; (c) a lower-confidence water-seepage candidate.}
\label{fig:qualitative}
\end{figure*}

Representative detections are shown in
Fig.~\ref{fig:qualitative}. The examples illustrate that the same
four-channel detector can locate both structurally distinct facade elements
and anomaly candidates within a single inference pass. The lower confidence
of the water-seepage example is consistent with its category-level result.

\subsection{Fourth-Channel Encoding Ablation}
\label{sec:encoding_ablation}

\begin{table}[t]
\centering
\caption{Fourth-channel encoding ablation on a single project-disjoint
split (M3T P4+P5 held-out fold, 130 images, 547 instances). This split and its
patience-100 schedule differ from the four-fold protocol of
Table~\ref{tab:project_detection}, so aggregate values are not comparable
across the two tables. Del.\ and Seep.\ denote $\mathrm{AP}_{50}$ for
delamination and water seepage.}
\label{tab:encoding_ablation}
\footnotesize
\setlength{\tabcolsep}{3.5pt}
\renewcommand{\arraystretch}{0.92}
\begin{tabular}{lrrrrrr}
\toprule
Fourth channel & P & R & $\mathrm{mAP}_{50}$ & $\mathrm{mAP}_{50:95}$
& Del. & Seep. \\
\midrule
None (RGB only)$^{\dagger}$ & 0.379 & 0.269 & \textbf{0.281} & \textbf{0.164}
& 0.000 & $<$0.001 \\
8-bit IR preview            & 0.577 & 0.193 & 0.206 & 0.112
& $<$0.001 & $<$0.001 \\
Linear 16-bit              & 0.585 & 0.220 & 0.235 & 0.129
& 0.021 & $<$0.001 \\
Signed $C_2$               & 0.560 & 0.273 & \textbf{0.281} & 0.145
& \textbf{0.073} & $<$0.001 \\
\bottomrule
\end{tabular}
\\[2pt]
{\footnotesize $^{\dagger}$The RGB-only variant uses a three-channel input
initialized from three-channel M3FD pretraining; the three four-channel
variants are initialized from four-channel M3FD pretraining. The channel
count is intrinsic to each variant and cannot be matched.}
\end{table}

Table~\ref{tab:encoding_ablation} isolates the effect of the fourth-channel
encoding under matched optimization on a single project-disjoint split. This
split and its patience-100 schedule differ from the four-fold protocol of
Table~\ref{tab:project_detection}, so the aggregate values are not directly
comparable across the two tables; the comparison instead separates the
contribution of the thermal encoding from that of the RGB input.

The RGB-only input does not detect delamination in this evaluation, reaching
an $\mathrm{AP}_{50}$ of 0.000. Under this configuration, delamination is
covered only by the variants that include a thermal channel. Among the three thermal encodings, the
signed local contrast $C_2$ gives the highest delamination
$\mathrm{AP}_{50}$, 0.073, against 0.021 for linear 16-bit normalization and
below 0.001 for the rendered 8-bit preview. The 8-bit and linear encodings
also reduce aggregate $\mathrm{mAP}_{50}$ relative to the RGB-only baseline,
to 0.206 and 0.235, whereas $C_2$ matches the RGB-only value of 0.281.
Aggregate $\mathrm{mAP}_{50:95}$ decreases from 0.164 to 0.145. Water seepage
stays below 0.001 for every variant and is not recovered by any thermal
encoding.

These results indicate that, in this facade setting, the thermal channel
contributes coverage of an anomaly category that the RGB-only detector does
not capture, rather than a uniform gain in aggregate accuracy, and that among
the evaluated encodings $C_2$ is the only one that adds this coverage without
lowering aggregate $\mathrm{mAP}_{50}$. They do not establish that RGB imagery
is in principle uninformative for delamination, only that the RGB-only
detector evaluated here does not recover it.

\subsection{Cross-Dataset Evaluation on RGBT-Tiny}
\label{sec:rgbt_tiny}

To examine portability beyond the M3T sensing regime, all four detectors
are trained on the corresponding RGBT-Tiny data and evaluated using their
best checkpoints~\cite{Ying2025RGBTTiny}. All methods receive paired visible
and thermal inputs; T and RGB denote the annotation source rather than the
input modality. The thermal frames are 8-bit single-channel JPEG images
instead of raw radiometric measurements. The proposed model therefore
constructs its fourth channel as an intensity-based local-contrast proxy,
rather than using the 16-bit radiometric $C_2$ representation employed on
M3T. This experiment evaluates the methods under an 8-bit RGB--thermal
setting rather than the benefit of raw radiometric sensing.

\begin{table}[t]
\centering
\caption{RGBT-Tiny results on loader-specific test supports. T and RGB
denote the annotation source.}
\label{tab:rgbt_tiny}
\footnotesize
\setlength{\tabcolsep}{2.3pt}
\renewcommand{\arraystretch}{0.92}
\begin{tabular}{@{}clrrrr@{}}
\toprule
Src. & Method & Images & Inst. &
$\mathrm{mAP}_{50}$ & $\mathrm{mAP}_{50:95}$ \\
\midrule
T & EME       & 12,590 & 189,089 & 0.497 & 0.249 \\
T & ICAFusion & 12,578 & 187,834 & 0.484 & 0.225 \\
T & CMA-Det   & 12,578 & 187,834 & 0.453 & 0.210 \\
T & Ours      & 12,590 & 189,089 & 0.167 & 0.060 \\
\midrule
RGB & EME       & 12,590 & 180,237 & 0.460 & 0.213 \\
RGB & ICAFusion & 12,576 & 178,752 & 0.469 & 0.208 \\
RGB & CMA-Det   & 12,576 & 178,752 & 0.151 & 0.049 \\
RGB & Ours      & 12,590 & 180,237 & 0.418 & 0.195 \\
\bottomrule
\end{tabular}
\end{table}

As shown in Table~\ref{tab:rgbt_tiny}, EME and the proposed model retain all
12,590 test pairs, whereas the ICAFusion and CMA-Det loaders retain 12,578
images for the T-source annotations and 12,576 for the RGB-source
annotations. The corresponding instance counts also differ; the results are
therefore treated as loader-specific rather than strictly sample-matched
comparisons. With T-source annotations, EME obtains 0.497/0.249 on the two
mAP measures, whereas the proposed detector obtains 0.167/0.060. With
RGB-source annotations, the proposed detector reaches 0.418/0.195, compared
with 0.460/0.213 for EME and 0.469/0.208 for ICAFusion.

The lower RGBT-Tiny scores should not be attributed to the thermal encoding
alone, because several factors change at once between the two settings.
RGBT-Tiny targets are typically small and are often characterized by their
overall thermal appearance against the sky, sea, or ground, whereas facade
targets are larger and lie on comparatively homogeneous surfaces.
Local-background subtraction may therefore attenuate the extended responses
on which RGBT-Tiny targets rely, while remaining suited to compact facade
deviations. Object scale, scene content, and annotation protocol also shift
together with the move from 16-bit radiometric data to rendered 8-bit frames.
These results are therefore most consistent with an intended operating regime
of aligned radiometric facade inspection rather than general-purpose 8-bit
thermal detection, and do not isolate the effect of the encoding by
themselves.

\section{Discussion}
\label{sec:discussion}

The experiments suggest that the usefulness of the radiometric contrast
channel may depend on the spatial extent of the thermal response. Delamination,
whose signature is comparatively compact, is the anomaly category for which
$C_2$ adds the clearest detection benefit, whereas the more diffuse
water-seepage response is not recovered by any evaluated encoding; a spatially
broad response may be partially absorbed into the locally estimated
background, which could limit the benefit of local contrast encoding for such
categories. The category-level results show a related distinction: windows,
air-conditioning units, and vents have stable geometric boundaries, whereas
surface cracks, efflorescence, delamination, and water seepage remain
difficult. Thermographic visibility is affected by surface material, thermal
excitation, and ambient
conditions~\cite{Balaras2002IRT,Kylili2014IRT,Mahmoodzadeh2023AerialThermography}.
The detector is therefore intended to identify candidate regions for
inspection, with final condition assessment retained as a human decision.

The landmark evaluation gives a median registration residual of
3.384 pixels on the sampled scenes. The controlled-displacement
analysis shows a decreasing in-box $C_2$ response for delamination, whereas the box-averaged water-seepage response does not decrease. This
limits the use of box-averaged contrast as a general measure of anomaly-signal
preservation. The scene-level homography was adequate for the present
acquisition geometry, but it assumes an approximately planar facade and a
stable sensor relationship within each inspection group. Its interactive
estimation also limits full automation. Calibrated extrinsics, automatic
cross-modal landmark estimation, and residual-quality checks could reduce
this dependence and reject poorly aligned pairs before detection.

The project-grouped comparison identifies two practical operating points.
ICAFusion achieves the highest mean $\mathrm{mAP}_{50:95}$ of 0.191, whereas
the proposed detector reaches 0.168 with 11.13 M parameters and 28.50 GFLOPs,
corresponding to approximately 91\% fewer parameters and 85\% fewer
floating-point operations. The RGBT-Tiny experiment further shows that the
representation is tied to its sensing regime: an intensity-based proxy
constructed from 8-bit thermal JPEG images does not reproduce the information
available in 16-bit radiometric measurements and may attenuate extended
object-level thermal appearance. The present M3T evaluation is also limited
to five projects with uneven category frequencies. Broader evaluation across
buildings, seasons, and sensor configurations, together with sample-matched
external testing, is therefore needed to characterize generalization.

\section{Conclusion}
\label{sec:conclusion}

This paper presented a sensor-level RGB--thermal pipeline for UAV facade
screening. The pipeline corrects the two sensors, maps RGB observations into
the thermal coordinate frame, retains their common image support, and encodes
16-bit radiometric measurements as a signed local contrast channel for a
simple four-channel single-stream detector. On M3T, the median residual
registration error is 3.384 pixels. Under
project-grouped four-fold evaluation, the detector obtains
$\mathrm{mAP}_{50}=0.294$ and $\mathrm{mAP}_{50:95}=0.168$ using 11.13 M
parameters and 28.50 GFLOPs. The external 8-bit RGBT-Tiny results further
clarify that the proposed representation is intended for aligned radiometric
facade inspection rather than unrestricted RGB--thermal object detection.
The resulting system provides a compact means of locating candidate facade
regions for subsequent expert review.

\bibliographystyle{IEEEtran}
\bibliography{references}

\end{document}